\documentclass{article} 

\usepackage{geometry}
\usepackage[square,sort,comma,numbers]{natbib}
\usepackage[parfill]{parskip}

\usepackage[utf8]{inputenc} 
\usepackage[T1]{fontenc}    
\usepackage{hyperref}       
\usepackage{url}            
\usepackage{booktabs}       
\usepackage{amsfonts}       
\usepackage{nicefrac}       
\usepackage{microtype}      
\usepackage{graphicx}
\usepackage{amsmath}
\usepackage{float}
\usepackage{booktabs}
\usepackage{tabularx}

\usepackage{enumitem}
\usepackage{xcolor}
\usepackage{algorithm}
\usepackage{algorithmic}
\usepackage{wrapfig}
\usepackage{bm}

\usepackage{bigstrut}
\makeatletter
\def\hlinewd#1{%
\noalign{\ifnum0=`}\fi\hrule \@height #1 %
\futurelet\reserved@a\@xhline}
\makeatother

\usepackage{xspace}
\newcommand{\gems}{\texttt{GEMS}\xspace}

\title{Model Aggregation via Good-Enough Model Spaces}

\author{%
    Neel Guha \\
    CMU\\
  \texttt{nguha@cmu.edu} \\
  \and
  Virginia Smith\\
  CMU \\
  \texttt{smithv@cmu.edu} \\
}

\begin{document}

\maketitle

\begin{abstract}
  In many applications, the training data for a machine learning task is partitioned across multiple nodes, and aggregating this data may be infeasible due to communication, privacy, or storage constraints. Existing distributed optimization methods for learning global models in these settings typically aggregate local updates from each node in an iterative fashion. However, these approaches require many rounds of communication between nodes, and assume that updates can be synchronously shared across a connected network. 
  In this work, we present Good-Enough Model Spaces (\gems)\footnote{Code available at \url{https://github.com/neelguha/gems}}, a novel framework for learning a global model by carefully intersecting the sets of "good-enough" models across each node. Our approach utilizes minimal communication and does not require sharing of data between nodes. We present methods for learning both convex models and neural networks within this framework and discuss how small samples of held-out data can be used for post-learning fine-tuning. In experiments on image and medical datasets, our approach on average improves upon other baseline aggregation techniques such as ensembling or model averaging by as much as 15 points (accuracy).
\end{abstract}

\section{Introduction}

There has been significant work in designing distributed optimization methods in
response to challenges arising from a wide range of large-scale learning applications. These  methods typically aim to train a global model by performing numerous communication rounds between distributed nodes. However, most approaches treat communication reduction as an objective, not a constraint, and seek to minimize the number of communication rounds while maintaining model performance. Less explored is the inverse setting---where the communication budget is fixed and we aim to maximize accuracy while restricting communication to only a few rounds. These \textit{few-shot} model aggregation methods can be beneficial when any of the following conditions hold:

\begin{itemize}[leftmargin=*]
    \item \textbf{Limited network infrastructure}: Distributed optimization methods typically require a connected network to support the synchronized collection of numerous learning updates (i.e. gradients). Such a network can be difficult to set up and maintain, especially in settings where nodes may represent different organizational entities (e.g., a network of different hospitals)~\cite{chang2018distributed}.  
    \item \textbf{Privacy and data ephemerality}: Privacy policies or regulations like GDPR may require nodes to periodically delete the raw local data~\cite{gdpr}. Few-shot methods enable learning an aggregate model in ephemeral settings, where a node may lose access to its raw data. Additionally, as fewer messages are sent between nodes, these methods have the potential to offer increased privacy benefits. 
    \item \textbf{Extreme asynchrony}: Even in settings where privacy is not a concern, messages from distributed nodes may be unevenly spaced and sporadically communicated over days, weeks, or even months (e.g., in the case of remote sensor networks). Few-shot methods drastically limit communication and thus  reduce the wall-clock time required to learn an aggregate model. 
\end{itemize}

Throughout this paper, we reference a simple motivating example.  Consider two hospitals, $\bm{A}$ and $\bm{B}$, which each maintain private (unshareable) patient data pertinent to some disease. As $\bm{A}$ and $\bm{B}$ are geographically distant, the patients they serve sometimes exhibit different symptoms. Without sharing the raw training data, $\bm{A}$ and $\bm{B}$ would like to jointly learn a single model capable of generalizing to a wide range of patients. The prevalent learning paradigm in this setting---distributed or federated optimization---dictates that $\bm{A}$ and $\bm{B}$   share iterative model updates (e.g., gradient information) over a network. This scenario can face many of the challenges described above, as hospitals operate under strict privacy practices, and may face legal, administrative, or ethical constraints prohibiting a shared connected network for training~\cite{chang2018distributed, mcclure2018distributed,sheller2018multi, tresp2016going}.

As a promising alternative, we present  \textit{Good-Enough Model Spaces} (\gems), a framework for learning an aggregate model over distributed nodes within a small number of communication rounds. Our framework draws inspiration from work in version space learning, an approach for characterizing the set of logical hypotheses consistent with available data~\cite{mitchell1978version}. Intuitively, the key idea in \gems is to leverage the fact that many possible hypotheses may yield `good enough' performance for a node's learning task on local data, and that considering the intersection between these sets of hypotheses (across nodes) can allow us to compute a global model quickly and easily. Despite the simplicity of the proposed approach, we find that it can significantly outperform other baselines such as ensembling or model averaging.

We make the following contributions in this work. First, we present a general formulation of the \gems framework. The framework itself is highly flexible, applying generally to black-box models and heterogeneous partitions of data. Second, we offer methods for calculating the `good-enough' model space on each node. Our methods are simple and interpretable in that each node only communicates its locally optimal model and a small amount of metadata corresponding to local performance. In the context of the above example, \gems serves to reduce the coordination costs for $\bm A$ and $\bm B$, enabling them to easily learn an aggregate model in a regime that does not require communicating incremental gradient updates. Finally, we empirically validate \gems on both standard image benchmarks (MNIST and CIFAR-10) as well as a domain-specific health dataset. We consider learning convex classifiers and neural networks in standard distributed settings, as well as scenarios in which some small global held-out data may be used for fine-tuning. On average, we find that \gems increases the accuracy of local baselines by 11 points and performs 43\% as well as a non-distributed model. With additional fine-tuning, \gems increases the accuracy of local baselines by 43 points and performs 88\% as well the non-distributed model. Our approach consistently outperforms other model aggregation baselines such as model averaging, and can either match or exceed the performance of ensemble methods with many fewer parameters.\\

\section{Related Work}
\textbf{Distributed Learning}. Current distributed and federated learning approaches typically rely on iterative optimization techniques to learn a global model, continually communicating updates between nodes until convergence is reached. To improve the overall runtime, a key goal in most distributed learning methods is to minimize communication for some fixed model performance. To this end, numerous methods have been proposed for communication-efficient and asynchronous distributed optimization~\citep[e.g.,][]{Large_scale_DL_dean_2012, Dekel:2012wm, param_server_Smola_14, fedavg, recht2011hogwild, shamir2014communication, COCOA_Smith_2016}. In this work, our goal is instead to maximize performance for a fixed communication budget (e.g., only one or possibly a few rounds of communication). 

\textbf{One-shot/Few-shot Methods}. While simple one-shot distributed communication schemes, such as model averaging, have been explored in convex settings~\cite{arjevani2015communication, mcdonald2009efficient,shamir2014communication, zhang2012communication,zinkevich2010parallelized}, guarantees typically rely on data being partitioned in an IID manner and over a small number of nodes relative to the total number of samples. Averaging can also perform arbitrarily poorly in non-convex settings, particularly when the local models converge to differing local optima~\cite{fedavg, sun2017ensemble}. Other one-shot schemes leverage ensemble methods, where an ensemble is constructed from models trained on distinct partitions of the data \cite{chawla2004learning,mcdonald2009efficient,sun2017ensemble}. While these ensembles can often yield good performance in terms of accuracy, a concern is that the resulting ensemble size can become quite large. In Section~\ref{sec:eval}, we compare against these one-shot baselines empirically, and find that \gems can outperform both simple averaging and ensembles methods while requiring significantly fewer parameters.

\textbf{Meta-learning, multi-task learning and transfer learning}. The goals of meta-learning, multi-task learning, and transfer learning are seemingly related, as these works aim to share knowledge from one learning process onto others. However, in the case of transfer learning, methods are typically concerned with one-way transfer---i.e., optimizing the performance of a single target model, not jointly aggregating knowledge between multiple models~\cite{pan2009survey}. In meta-learning and multi-task learning, such joint optimization is performed, but similar to traditional distributed optimization methods, it is assumed that these models can be updated in an iterative fashion, with potentially numerous rounds of communication performed throughout the training process ~\cite{evgeniou2004regularized, DBLP:journals/corr/FinnAL17}.

\textbf{Version Spaces}. In developing \gems, we draw inspiration from work in \textit{version space learning}, an approach for characterizing the set of logical hypotheses consistent with available data~\cite{mitchell1978version}. Similar to~\cite{balcan2012distributed}, we observe that if each node communicates its version space to the central server, the server can return a consistent hypothesis in the intersection of all node version spaces. However, \cite{balcan2012distributed, mitchell1978version} assume that the hypotheses of interest are consistent with the observed data, i.e., they perfectly predict the correct outcomes. Our approach significantly generalizes this notion to explore imperfect, noisy hypotheses spaces as more commonly observed in practice.

\section{Good-Enough Model Spaces (\gems)}
As in traditional distributed learning, we assume a training set $S = \{(x_i, y_i)\}^m_{i=1}$ drawn from $\mathcal{D}_{\mathcal{X} \times \mathcal{Y}}$ is  divided amongst $K$ nodes\footnote{We use `node' to refer abstractly to distributed entities such as devices, machines, organizations, etc.} in a potentially non-IID fashion. We define  $S^k := \{(x^k_1, y^k_1), ... \}$ as the subset of training examples belonging to node $k$, such that $\sum_{k=1}^K \vert S^k \vert = m$, and assume that a single node (e.g., a central server) can aggregate updates communicated in the network. Fixing a function class $\mathcal{H}$, our goal is to learn an aggregate model $h_G \in \mathcal{H}$ that approximates the performance of the optimal model $h^*\in \mathcal{H}$ over $S$ while limiting communication to one (or possibly a few) rounds of communication. 

In developing a method for model aggregation, our intuition is that the aggregate model should be at least \textit{good-enough} over each node's local data, i.e., it should achieve some minimum performance for the task at hand. Thus, we can compute $h_G$ by having each node compute and communicate a \textit{set} of locally good-enough models to a central server, which learns $h_G$ from the intersection of these sets. 

Formally, let $Q: (\mathcal{H}, \{(x_i, y_i)\}^d ) \rightarrow \{-1,1\}$ denote a \textit{model evaluation function}, which determines whether a given model $h$ is good-enough over a sample of $d$ data points $\{(x_i, y_i)\}^d \subseteq S$. In this work, define "good-enough" in terms of the accuracy of the model $h$ and a threshold $\epsilon$:

\begin{equation}
  Q(h, \{(x_i, y_i)\}^d) =
  \begin{cases}
	1 & \dfrac{1}{d} \sum_{i=1}^{d} \mathbb{I}\{h(x_i) = y_i\} \geq \epsilon \\
	-1 & \text{else} 
  \end{cases}
  \label{eq:modeleval}
\end{equation}

Using these model evaluation functions, we formalize the proposed approach for model aggregation, \gems, in Algorithm \ref{alg:GEMS}. In \gems, each node $k = 1,...,K$ computes the set of models $H_k = \{h_1,...,h_n|h_i \in \mathcal{H}, Q_k(h_i, S^k) = 1\}$ and sends this set to the central node. After collecting $H_1,...H_K$, the central node selects $h_G$ from the intersection of the sets, $\cap_k H_k$. When granted access to a small sample of public data, the server can additionally use this auxiliary data further fine-tune the selected $h$ $\in$ $\cap_k H_k$, an approach we discuss in Section \ref{fine-tuning}.

 \begin{figure}
 \begin{center}
 \vspace{-10mm}
 \includegraphics[scale=0.4]{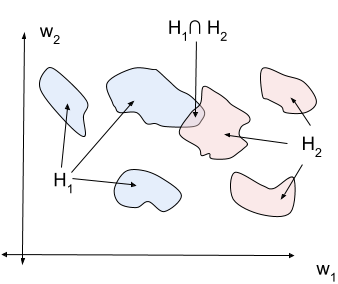}
 \end{center}
 \caption{Depiction of \gems}
 \label{eps_ex}
 \end{figure}

Figure \ref{eps_ex} visualizes this approach for a model class with only two weights ($w_1$ and $w_2$) and two nodes ("red" and "blue"). The `good-enough' model space, $H_k$, for each node is a set of regions over the weight space (the blue regions correspond to one node and the red regions correspond to second node). The final aggregate model, $h_G$, is selected from the area in which the spaces intersect. Section \ref{section:intersection} empirically analyzes when intersections may not exist. In practice, finding an intersection depends on (1) how $\epsilon$ is set for each node, and (2) the complexity of the task with respect to the model. 

For a fixed hypothesis class $\mathcal{H}$, applying Algorithm \ref{alg:GEMS} requires two components: (i) a mechanism for computing $H_k$ over every node, and (ii) a mechanism for identifying the aggregate model, $h_G \in \cap_{k} H_k$. In this work, we present such methods for two types of models: convex  models (Section \ref{convex}) and neural networks (Section \ref{non-convex}). For convex models, we find that $H_k$ can be approximated as $\mathbb{R}^d$-ball in the parameter space, requiring only a single round of communication between nodes to learn $h_G$. For neural networks, we apply Algorithm \ref{alg:GEMS} to each layer in a step-wise fashion, and compute $H_k$ as a set of independent $\mathbb{R}^d$-balls corresponding to every neuron in the layer. This requires one round of communication per layer (a few rounds for the entire network).

\begin{algorithm}
\caption{\gems Meta-Algorithm}
\label{alg:GEMS}
\begin{algorithmic}[1]
    \STATE {\bf Input:}  $S = \{(x_i, y_i)\}^m_{i=1}$
    \FOR  {$k=1, \cdots, m$ in parallel}
        \STATE Node $k$ computes good-enough model space, $H_k$, according to~\eqref{eq:modeleval}
    \ENDFOR
    \STATE Return intersection $h_G \in \cap_{k} H_k$
\end{algorithmic}
\end{algorithm}
\vspace{-4mm}

\subsection{Convex Classifiers}\label{convex}
Consider a convex classifier $f_w(\cdot)$ parameterized by a weight vector $w \in \mathbb{R}^d$. Below, we describe the two key steps of \gems (Algorithm~\ref{alg:GEMS}) for such classifiers, including construction of the good-enough models, $H_k$,  and computation of the intersection, $h_G$. 

For each node $k$, we can compute the set of good-enough models, $H_k$, as an $\mathbb{R}^d$-ball in the parameter space, represented as a tuple $(c_k \in \mathbb{R}^d, r_k\in\mathbb{R})$ corresponding to the center and radius. Formally, $H_k = \{w \in \mathbb{R}^d| ||c_k - w||_2 \leq r_k\}$. While this is a simple notion of a model space, we find that it works well in practice (Section~\ref{sec:eval}), and has the added benefit of reducing communication costs, as each node only needs to share its center $c_k$ and radius $r_k$. Fixing $\epsilon$ as our minimum acceptable performance, we want to compute $H_k$ according to~\eqref{eq:modeleval} such that $\forall w \in H_k$, $Q(w, S^k) = 1$. In other words, every model contained within the $d$-ball should have an accuracy greater than or equal to $\epsilon$.

\textbf{Construction.} Algorithm \ref{alg:h_k} presents the $H_k$ construction algorithm for node $k$ and data $S^k$, where 
$$w^*_k = \arg\min_w \dfrac{1}{|S^k|}\sum_{i=1}^{|S^k|} \ell(f_w(x_i), y_i) \, ,$$ 
$\epsilon$ is fixed a hyperparameter, and $Q(\cdot)$ is a minimum accuracy threshold defined according to~\eqref{eq:modeleval}. In Algorithm~\ref{alg:h_k}, $R_\text{max}$ and $\Delta$ define the scope and stopping criteria for the binary search.  Intuitively, Algorithm 2 approximates the maximum radius for an $\mathbb{R}^d$ ball centered at $w^*_k$ such that all points contained within the ball correspond to 'good-enough' models (defined by $Q(\cdot)$).

\textbf{Intersection.} After constructing the sets of good-enough models, $H_k$, \gems computes an aggregate model in the intersection $h_G\in\cap_kH_k$. Given $K$ nodes with individual model spaces $H_k = (c_k, r_k)$, we can pick a point in this intersection by solving:
\begin{equation}
h_G = \arg\min_w \sum_{k=1}^K \max(0, ||c_k - w||_2 - r_k) 
\label{eq:convexint}
\end{equation}
which takes a minimum value of $0$ when $w \in \cap_k H_k$. In practice, we solve ~\ref{eq:convexint} via gradient descent. As we discuss in Section \ref{fine-tuning}, this $w$ can be improved by fine-tuning on a limited sample of publicly available data. In practice, we observe that this simple $\mathbb{R}^d$ ball construction can also be  extended to ellipsoids by letting the radius vary across different dimensions. The minor modifications required for this are discussed in Appendix \ref{appen:ellipsoid}.
\begin{algorithm}
\caption{ConstructBall}
\label{alg:h_k}
\begin{algorithmic}[1]
    \STATE {\bf Input:}  $k$, $f_w(\cdot)$, $Q(\cdot)$,$w^*_k$ $S^k = \{(x^k_1, y^k_1),..\}$, $R_\text{max}$, $\Delta$
    \STATE Sets $c_k$ to $w^*_k$.
    \STATE Initialize $R_\text{lower} = 0$, $R_\text{upper} = R_\text{max}$
    \WHILE {$ R_\text{upper} - R_\text{lower} > \Delta$}
        \STATE Set $R = \text{Avg}(R_\text{upper}, R_\text{lower})$
        \STATE Sample $w_1,...,w_p$ from surface of $\mathbb{B}_R(c_k)$
        \IF{$Q(f_{w'}, S^k) = 1, \forall w' = w_1, ..., w_p'$}
            \STATE Set $R_\text{lower} = R$
        \ELSE
            \STATE Set $R_\text{upper} = R$
        \ENDIF
    \ENDWHILE
    \STATE Return $H_k$
\end{algorithmic}
\end{algorithm}

\subsection{Neural Networks}\label{non-convex}
We additionally consider \gems with a simple class of neural networks, multi-layer perceptrons (MLPs). 
 First, we observe that the final layer of an MLP is a linear model. Hence, we can apply the method above with no modification. However, the input to this final layer is a set of stacked, non-linear transformations which extract feature from the data. For these layers, the approach presented above faces two challenges: 
 \begin{enumerate}[leftmargin=*]
     \item \textbf{Node specific features}: When the distribution of data is non-IID across nodes, different nodes may learn different feature extractors in lower layers. 
     \item \textbf{Model isomorphisms}: It is well-known that MLPs are highly sensitive to weight initialization~\cite{glorot2010understanding}. Two models trained on the same set of samples (with different initializations) may have equivalent behavior despite learning different weights. In particular, reordering a model's hidden neurons (within the same layer) may not alter the model's predictions, but corresponds to a different weight vector $w$.  
 \end{enumerate}
 
 In order to construct $H_k$ for hidden layers, we modify the approach presented in Section \ref{convex}, applying it to individual hidden neurons. Formally, let the ordered set $[f^j_{w_1}(\cdot),...,[f^j_{w_L}(\cdot)]$ correspond to the set of $L$ hidden neurons in layer $j$. Here, $f^j_{w_l}(\cdot) = g(w_l^T z^{j-1})$ denotes the function computed by the $l$-th neuron over the output from the previous layer $z^{j-1}$, with $g(\cdot)$ corresponding to some non-linearity (e.g. ReLU). Fixing an indexed ordering over $d$ data points, let $z^j_l = [(z^j_l)_1, ..., (z^j_l)_d]$ denote the vector of activations produced by $f^j_{w_l}(\cdot)$. Similar to the model evaluation function, $Q$ in~\eqref{eq:modeleval}, we can define an alternative $Q$ over a neuron in terms of $z^{j-1}$ and $z^j_l$ (the neuron's input and output): 
 
 \begin{equation}
     Q_\text{neuron}(w', \{((z^{j-1})_i, (z^{j}_l)_i)\}^d) =
  \begin{cases}
	1 & \dfrac{1}{d} \sqrt{\sum_{i=1}^{d} \big(f^j_{w'}(z^{j-1})_i) - (z^{j}_l)_i\big)^2} \leq \epsilon_j \\
	-1 & \text{else} 
  \end{cases}
  \label{eq:neuroneval}
 \end{equation}
 
 Broadly, $Q_\text{neuron}$ returns $1$ if the output of $f_{w'}$ over $z^{j-1}$ is within $\epsilon$ of $z^j_l$, and $-1$ otherwise. Using this neuron-specific evaluation function, we can now apply Algorithm \ref{alg:h_k} to each neuron. Formally: 
 \begin{enumerate}[leftmargin=*]
    \item Each node $k$ learns a locally optimal model $m^k$, with optimal neuron weights ${w^j_l}^*$, over all $j$, $l$.
    \item Fix hidden layer $j=1$. Apply Algorithm \ref{alg:h_k} to each hidden neuron $[f^j_{w_1}(\cdot),...,[f^j_{w_L}(\cdot)]$, with $Q(\cdot)$ according to~\eqref{eq:neuroneval} with hyperparameter $\epsilon_j$. Denote the $\mathbb{R}^d$ ball constructed for neuron $l$ as $H^{k}_{j, l}$.
    \item Each node communicates its set $H^k_{j, \cdot} = [H^{k}_{j, 1}, ...,H^{k}_{j, L}]$ to the central server which constructs the aggregate hidden layer ${f_G}_{j,\cdot}$ such $\forall i, k, \exists i': {f_G}_{j,i'} \in H^k_{j, i}$. This is achieved by greedily applying~\eqref{eq:convexint} to tuples in the cartesian product $H^1_{j, \cdot}\times...\times H^K_{j, \cdot}$. Neurons for which no intersection exists are included in ${f_G}_{j,\cdot}$, thus trivially ensuring the condition above.
    \item The server sends ${h_G}_{j,\cdot}$ to each node, which insert ${h_G}_{j,\cdot}$ at layer $j$ in their local models and retrain the layers above $j$. 
 \end{enumerate}

\begin{figure}
    \vspace{-20pt}
    \centering
    \includegraphics[scale=0.3]{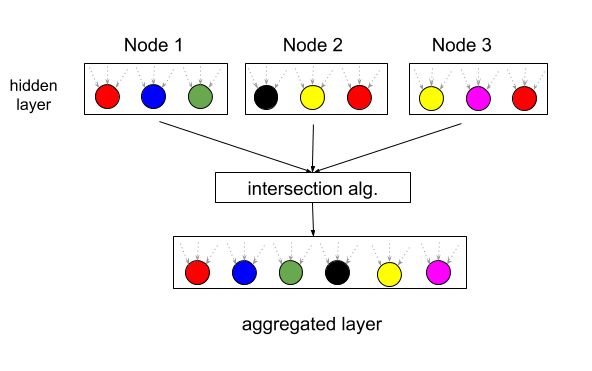}
    \caption{In \gems, the aggregate hidden layer is composed of intersections between different neurons from node local models.}
    \label{fig:gems_dnn}
 \end{figure}

Steps (1)-(4) can be repeated until no hidden layers remain. We note that step (3) can be expensive when there are a large number of hidden neurons $L$ and nodes $K$, as $|H^1_{j, \cdot}\times...\times H^K_{j, \cdot}|$ increases exponentially. A simplifying assumption is that if $H^k_{j, i}$ and  $H^k_{j, l}$ are `far', then the likelihood of intersection is low. Using this intuition, we can make the method more scalable by performing k-means clustering over all neurons. In step (3), we now only look for intersections between tuples of neurons in the same cluster. Neurons for which no intersection exists are included in ${f_G}_{j,\cdot}$. We explore this procedure empirically in Section~\ref{sec:eval}. For notational clarity, we denote the number of clusters with which k-means is run as $m_\epsilon$, in order to distinguish it from node index $k$. Figure \ref{fig:gems_dnn} displays this intersection procedure. Neurons with intersecting $H_j^k$ are denoted by the same color.
 
 \subsection{Fine-tuning}
 \label{fine-tuning}
 
 In many contexts, a small sample of public data, $S_\text{public}$, may be available to the central server. For example, this may correspond to a public research dataset, or to nodes that have waived their privacy right for some data. In these scenarios, the coordinating server can fine-tune $H_G$ on $S_\text{public}$ by updating the weights for a small number of epochs. In Section \ref{section:convex_results} and \ref{section:dnn_results}, we find that fine-tuning is particularly useful for improving the quality of the \gems model, $h_G$, compared to other baselines.  

 \section{Evaluation}
 \label{sec:eval}

 We now present the empirical results to validate the effectiveness of \gems. In Section \ref{section:overview}, we briefly describe our experimental setup. In Section \ref{section:convex_results} we apply \gems to  logistic regression models, and in Section \ref{section:dnn_results} we apply \gems to two layer neural networks. In Section \ref{section:fine}, we demonstrate that fine-tuning is particularly beneficial for the \gems model compared to other baselines. Finally, we discuss the effect of $\epsilon$ on model size in Sections \ref{section:model_size}.
 
 \subsection{Experimental Setup}\label{section:overview}

 \textbf{Datasets.} We explore results on three datasets: MNIST \cite{mnist}, CIFAR-10 \cite{krizhevsky2009learning}, and HAM10000 \cite{DBLP:journals/corr/abs-1803-10417}, a medical imaging dataset. HAM10000 (HAM) consists of images of skin lesions, and our model is tasked with distinguishing between $7$ types of lesions. Full details on all datasets can be found in Appendix \ref{appen:preproc}.
 We split all three datasets into disjoint train, test, and validation splits. Train and validation sets were then further partitioned across different nodes. All results are reported on the test set. To demonstrate the effectiveness of our approach in difficult heterogeneous settings, we partitioned data by label, such that all train/validation images corresponding to a particular label would be assigned to the same node. Full details on  partitioning can be found in Appendix \ref{appen:data_split}.

 \textbf{Baselines.} In Section \ref{section:convex_results} and \ref{section:dnn_results}, we evaluate against the following single-model baselines: 
 \vspace{-2mm}
 \begin{itemize}[leftmargin=*]
   \setlength\itemsep{-0.0em}
     \item \textbf{Global}: A model trained on data aggregated across all nodes. This is an unachievable ideal, as it requires nodes to share data.
     \item \textbf{Local}: A model trained locally on a node, with only data belonging to that node. When reporting the global performance, we average the performance of each node's local model. 
     \item \textbf{Naive Average}: A model produced by averaging the parameters of each local model. 
 \end{itemize}
  \vspace{-2mm}
 In Section \ref{section:model_size}, we evaluate against an ensemble baseline. We create a majority-vote ensemble from node local models, with random selection for ties. We find that \gems is more parameter efficient, delivering better performance with few parameters.

 For \gems, we present two results. First, we report the accuracy of a model learned applying either the convex or non-convex variant of Algorithm  \ref{alg:h_k}. Next, we also report the accuracy after fine-tuning this model on a small sample of public data (described in more detail below). Each node computes $H_k$ over the local node validation split. We report the average accuracy (and standard deviation) of all results over $5$ trials. Randomness across trials stems from weight initialization---i.e., each node's local model is initialized with different weights in each trial.

 \subsection{Convex Classifiers}\label{section:convex_results}
We evaluate the convex variant of \gems on logistic regression classifiers. The results for all three datasets for over $5$ nodes are presented in Table \ref{table:convex}. Fine-tuning consists of updating the weights of the \gems model for 5 epochs over a random sample of 1000 images from the aggregated validation data. Training details are provided in Appendix \ref{appen:convex}.

Results in Table \ref{table:convex} correspond to the ellipsoidal variant of \gems, where the relative ratios of radii axes are computed in proportion to each parameter's Fisher information (Appendix \ref{appen:ellipsoid}). \gems comprehensively outperforms both local models and naive averaging. Fine-tuned \gems significantly outperforms all baselines, and comes relatively close to the global model. We use $\epsilon = 0.40$ for MNIST, $\epsilon = 0.20$ for HAM, and $\epsilon = 0.20$ for CIFAR-10 (over $5$ nodes). Complete results over additional agent partitions are described in Appendix \ref{appen:convex_results}.

 \begin{table}[h]
 \renewcommand{\arraystretch}{1.4}
    \caption{Convex Results ($K = 5$) }
    \label{table:convex}
    \begin{tabular}{l|c|ccc|c}
     \hlinewd{1pt}
    \bf Dataset &  \bf Global & \bf Local &\bf Averaged  &\bf \gems &\bf \gems Tuned  \\ \hlinewd{0.5pt}
    MNIST &  0.926 (0.001)&	0.198 (0.010)&	0.444 (0.028)&	\bf 0.456 (0.020)&	0.877 (0.005)     \\ \hlinewd{0.5pt}
    CIFAR-10  & 0.597 (0.007)&	0.178 (0.010)  &	0.154 (0.024)&	\bf 0.210 (0.020)&	0.499 (0.016)  \\ \hlinewd{0.5pt}
    HAM    & 0.559 (0.002)&	0.237 (0.050)&	0.185 (0.002)&\bf	0.348 (0.048)&	0.530 (0.009)  \\ \hlinewd{0.75pt}
    \end{tabular}
\end{table}

 \subsection{Neural Networks}\label{section:dnn_results}
 We evaluate the non-convex variant of \gems on simple two layer feedforward neural networks (Table \ref{table:dnn}). The precise network configuration and training details are outlined in Appendix~\ref{appen:nonconvex}. Fine-tuning consists of updating the last layer's weights of the \gems model for 5 epochs over a random sample of 1000 images from the aggregated validation data. In the majority of cases, the untuned \gems model outperforms the local/average baselines. Fine-tuning has a significant impact, and fine-tuned \gems outperforms the local/average baselines. Additional results are provided in Appendix \ref{appen:non_convex_results}.

  \begin{table}[h]
  \renewcommand{\arraystretch}{1.4}
    \caption{NN Results ($K = 5$)}
    \label{table:dnn}
    \begin{tabular}{l|c|ccc|c}
    \hlinewd{1pt}
    \bf Dataset &  \bf Global    & \bf Local     &\bf Averaged      &\bf \gems       &\bf \gems Tuned     \\ \hlinewd{0.5pt}
        MNIST      & 0.965 (0.001)  & 0.199 (0.010) & 0.259 (0.039) &  \bf 0.439 (0.044) & 0.886 (0.007)     \\ \hlinewd{0.5pt}
        CIFAR-10  & 0.651 (0.004)  & 0.183 (0.009) & 0.128 (0.023) &  \bf 0.223 (0.011) & 0.502 (0.011)    \\ \hlinewd{0.5pt}
        HAM &  0.606 (0.006) &	\bf0.250 (0.048)	&0.153 (0.018)	&0.190 (0.002)	&0.544 (0.017)     \\ \hlinewd{0.75pt}
\end{tabular}
\end{table}

 \subsection{Fine-tuning}\label{section:fine}
 The results in Table \ref{table:convex} and \ref{table:dnn} suggest that fine-tuning can have a significant effect on the \gems model. In this section, we demonstrate that the benefit of fine-tuning is disproportionate: fine-tuned \gems outperforms the fine-tuned baselines. We apply the same fine-tuning technique to each node's local model (\textbf{fine-tuned local}) and the parameter average of all node local models (\textbf{fine-tuned average}). In addition, we compare to a model trained solely on the public sample used for fine-tuning (\textbf{raw}).

\begin{figure}
 \centering
\includegraphics[width=.33\textwidth]{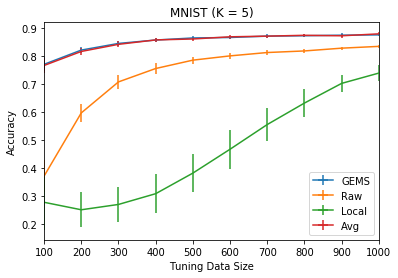}\hfill
\includegraphics[width=.33\textwidth]{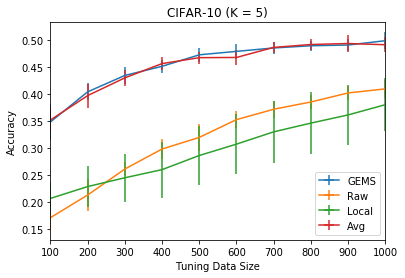}\hfill
\includegraphics[width=.33\textwidth]{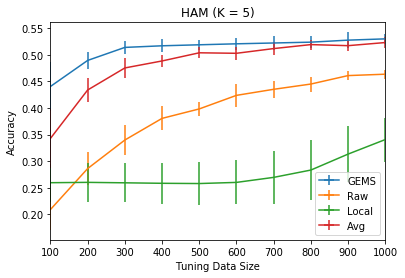}
\caption{Comparative effects of fine-tuning for \gems vs Baselines (Convex)}
\label{fig:logit_fine}
 \end{figure}

\begin{figure}
 \centering
\includegraphics[width=.33\textwidth]{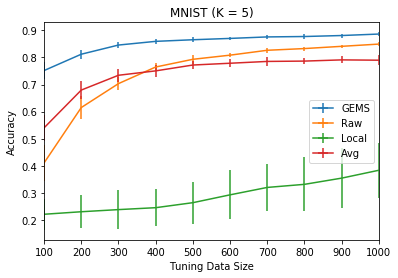}\hfill
\includegraphics[width=.33\textwidth]{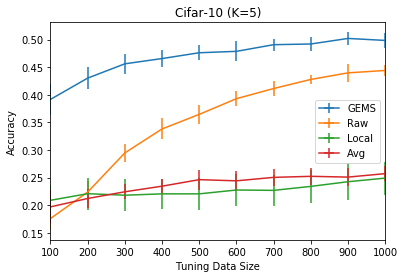}\hfill
\includegraphics[width=.33\textwidth]{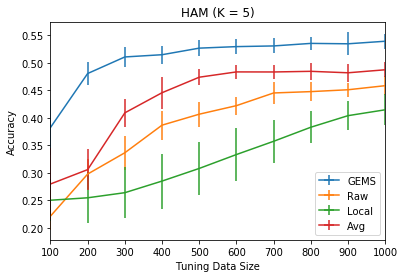}
\caption{Comparative effects of fine-tuning for \gems vs Baselines (Neural Network)}
\label{fig:dnn_fine}
 \end{figure}

 We can evaluate the effect of fine-tuning as the number of public data samples (the size of the tuning set) changes. In convex settings (Figure \ref{fig:logit_fine}), fine-tuned \gems performs comparably to the fine-tuned average model, and both outperform fine-tuning the other baselines. In non-convex settings (Figure \ref{fig:dnn_fine}), fine-tuned \gems consistently outperforms the fine-tuned baselines, regardless of the sample size. This suggest that the \gems model is learning weights that are more amenable to fine-tuning, and are perhaps able to capture better representations for the overall task. Though this advantage diminishes as the tuning sample size increases, the advantage of \gems is especially pronounced for smaller samples. With just 100 public data samples, \gems achieves an average improvement (accuracy) of 22.6 points over raw training, 15.1 points over the fine-tuned average model, and 26.8 points over the fine-tuned local models.

 \subsection{Model Size}\label{section:model_size}
 In non-convex settings, \gems provides a modular framework to tradeoff between the model size and performance, via hyperparameters $m_\epsilon$ (the number of clusters created when identifying intersections) and $\epsilon_j$ (the maximum output deviation allowed for hidden neurons). Intuitively, both parameters control the number of hidden neurons in the aggregate model $h_G$. Table \ref{table:model_size} compares adjustments for $\epsilon_j$ and $m_\epsilon$ on CIFAR-10 for 5 nodes against an ensemble of local node models. We observe that the \gems performance correlates with the number of hidden neurons, and that \gems outperforms the ensemble method at all settings (despite having fewer parameters). We observe a similar trend for HAM and MNIST, which are presented in more detail in Appendix \ref{appen:ensemble}.
 
 \begin{table}[h]
 \centering
 \caption{Model Size Results (CIFAR-10, $K= 5$)}
    \label{table:model_size}
\begin{tabular}{lcc}
\toprule
\bf Method                                              & \bf Accuracy      & \bf \# hidden neurons \\ \midrule
Tuned \gems ($m_\epsilon = 150$, $\epsilon_j = 0.7$) & 0.454 (0.018) & 163.40 (1.20)            \\ \midrule
Tuned \gems ($m_\epsilon = 150$, $\epsilon_j = 0.5$) & 0.492 (0.012) & 246.20 (8.93)            \\ \midrule
Tuned \gems ($m_\epsilon = 200$, $\epsilon_j = 0.3$) & 0.502 (0.011) & 379.60 (6.68)            \\ \midrule
Tuned \gems ($m_\epsilon = 100$, $\epsilon_j = 0.3$) & 0.501 (0.011) & 386.00 (18.76)           \\ \midrule
Ensemble                                            & 0.194 (0.005) & 500.00 (0.0)            \\ \bottomrule
\end{tabular}
\end{table}

\subsection{Intersection Analysis}\label{section:intersection}
A natural question to ask in \gems is how sensitive the results are to the hyperparameter, $\epsilon$. In certain cases, \gems may not find an intersection between different nodes when the task is too complex for the model, or $\epsilon$ is set too high. In practice, we observe that finding an intersection requires being conservative (i.e., low values) when setting $\epsilon$ for each node. We explain this by our choice to represent $H_k$ as a $\mathbb{R}^d$ ball. Although $\mathbb{R}^d$ balls are easy to compute and intersect, they may be fairly coarse approximations of the actual good-enough model space. This suggests that the current results may be improved even further by considering methods for computing/intersecting more complex model spaces. To illustrate node behavior at different settings of $\epsilon$, we defer the reader to experiments performed in Appendix \ref{appen:intersect}.

\section{Conclusion}
In summary, we introduce good-enough model spaces (\gems), an intuitive framework for learning an aggregate model across distributed nodes within one round of communication (for convex models) or a few rounds of communication (for neural networks). Our method is both simple and effective, requiring that nodes only communicate their locally optimal models and corresponding good-enough model space. We achieve promising results with relatively simple approximations of the good-enough model space and on average achieve 88\% of the accuracy of the ideal non-distributed model. In future work, we intend to explore more complex representations of the good-enough model space and believe these could significantly improve performance. 

 \section{Acknowledgements}
 
 We thank Tian Li, Michael Kuchnik, Anit Kumar Sahu, Otilia Stretcu, and Yoram Singer for their helpful comments. This work was supported in part by the National Science Foundation grant IIS1838017, a Google Faculty Award, a Carnegie Bosch Institute Research Award, and the CONIX Research Center. Any opinions, findings, and conclusions or recommendations expressed in this material are those of the author(s) and do not necessarily reflect the National Science Foundation or any other funding agency.

\bibliographystyle{abbrv}
\bibliography{sources}
\newpage
\appendix
\section{Non-Uniform $\mathbb{R}^d$ Balls}\label{appen:ellipsoid}
Algorithm 2 approximates the good-enough model space on node $k$ as  $\mathbb{B}_R(w_k^*)$, where $w_k^*$ corresponds to the locally optimal weights on node $k$ and $R$ is estimated on local validation data. An $\mathbb{R}^d$ ball is a coarse approximation of the good-enough model space, and assumes that good-enough model space is equally sensitive to perturbations over all parameters. 

A reasonable alternative is to model the good-enough model space as an ellipsoid, which is more sensitive to perturbations on certain parameters. Fixing axis radii $r_1, ..., r_d$, the good-enough model space is given by: 
\begin{equation}
    H_k = \bigg\{w \in \mathbb{R}^d \bigg|\sum_{i=1}^d \dfrac{(w_i - {w_k^*}_i)^2}{r_i^2} \leq 1 \bigg\}
\end{equation}
The larger the radii for a particular parameter, the more variation is allowed (while still remaining in $H_k$). Thus, $r_i$ should implicitly capture the effect of $w_i$ on the model's output (i.e. the \textit{sensitivity} of $w_i$). In practice, this can be computed by the inverse Fisher information~\cite{DBLP:journals/corr/KirkpatrickPRVD16}. For computed Fisher information values $F_1,\dots, F_d$ over ${w_k^*}_1,\dots {w_k^*}_d$:
\begin{equation}
    r_i = \max\bigg(\dfrac{\min_j F_j}{F_i}, c\bigg)\cdot R
\end{equation}
where $0 < c < 1$. This forces $cR\leq r_i \leq R$, guaranteeing that the radius for the most sensitive parameter (i.e. with the highest Fisher information) scales by a constant factor $c$ with the radius of the least sensitive parameter. In practice, Algorithm 2 can now be applied to approximate $R$, where in line (6) models are sampled from the surface of an ellipsoid with radii $r_1\cdot R, \dots, r_d\cdot R$. When all parameters $w_i$ are equally sensitive, i.e. $F_1 = F_2 = \dots F_d$, then $H_k$ generalizes to an $\mathbb{R}^d$ ball with radius $R$.  

For the convex results in Section \ref{section:convex_results}, we find that Fisher information based ellipsoids outperform fixed radius $\mathbb{R}^d$ balls and report ellipsoidal results. 

\section{Experimental Setup}\label{appen:exp_overview}

\subsection{Preprocessing}\label{appen:preproc}
We describe preprocessing/featurization steps for our empirical results. \\

\textbf{MNIST}. We used the standard MNIST dataset.

\textbf{CIFAR-10}. We featurize CIFAR-10 (train, test, and validation sets) using a pretrained ImageNet VGG-16 model \cite{simonyan2014very} from Keras. All models are learned on these featurized images. 

\begin{figure}[h]
    \centering
    \includegraphics[scale=0.5]{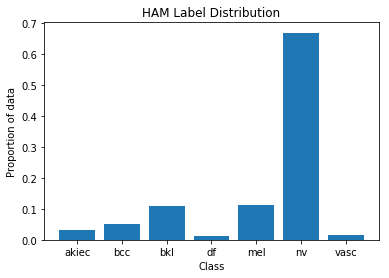}
    \caption{Distribution of classes for HAM}
    \label{fig:ham_dist}
\end{figure}

\textbf{HAM10000}. The HAM dataset consists of $10015$ images of skin lesions. Lesions are classified as one of seven potential types: actinic keratoses and intraepithelial carcinoma (akiec), basal cell carcinoma (bcc), benign keratosis (bkl), dermatofibroma (df), melanoma (mel), melanocytic nevi (nv), and vascular lesions (vasc). As Figure \ref{fig:ham_dist} shows, the original original dataset is highly skewed, with almost 66\% of images belonging to one class. In order to balance the dataset, we augment each class by performing a series of random transformations (rotations, width shifts, height shifts, vertical flips, and horizontal flips) via Keras \cite{chollet2015keras}. We sample 2000 images from each class. We initially experimented with extracting ImageNet features (similar to our proceedure for CIFAR-10). However, training a model on these extractions resulted in poor performance. We constructed our own feature extractor, by training a simple convolutional network on 66\% of the data, and trimming the final 2 dense layers. This network contained 3 convolutional layers ($32$, $64$, $128$ filters with $3\times 3$ kernels) interspersed with $2\times 2$ MaxPool layers, and followed by a single hidden layer with $512$ neurons. 

\subsection{Data Partitioning}\label{appen:data_split}
Given $K$ nodes, we partitioned each dataset in order to ensure that all images corresponding to the same class belonged to the same node. Table \ref{table:lab_part} provides an explicit breakdown of the label partitions for each of the three datasets, across the different values of $K$ we experimented with. Where possible, we partition data such that each node receives all images associated with a unique subset of the labels. An exception is made for HAM over $5$ nodes, where labels $0-4$ are each assigned to unique nodes, and labels $5$ and $6$ are uniformly divided amongst the $5$ nodes. Each node's images are always unique - no image is replicated across multiple nodes.    

\begin{table}[h]
\centering
\begin{tabular}{lll}\toprule
    \bf Dataset & Number of nodes ($K$)   & \bf Label Division \\ \midrule
    MNIST       & 2     & $[\{0, 1, 2, 3, 4\}, \{0, 1, 2, 3, 4\}]$ \\ \midrule
    MNIST       & 3     & $[\{0, 1, 2\},\{3, 4, 5\}, \{6, 7, 8, 9\}]$ \\ \midrule
    MNIST       & 5     & $[\{0, 1\},\{2, 3\}, \{4, 5\}, \{6, 7\}, \{8, 9\}]$ \\ \midrule
    CIFAR10     & 2     & $[\{0, 1, 2, 3, 4\}, \{0, 1, 2, 3, 4\}]$ \\ \midrule
    CIFAR10     & 3     & $[\{0, 1, 2\},\{3, 4, 5\}, \{6, 7, 8, 9\}]$ \\ \midrule
    CIFAR10     & 5     & $[\{0, 1\},\{2, 3\}, \{4, 5\}, \{6, 7\}, \{8, 9\}]$ \\ \midrule
    HAM         & 2     & $[\{0, 1, 2, 3\}, \{4, 5, 6\}]$ \\ \midrule
    HAM         & 3     & $[\{0, 1\}, \{2, 3\}, \{4, 5, 6\}]$ \\ \midrule
    HAM         & 5     & $[\{0, 5, 6\}, \{1, 5, 6\}, \{2, 5, 6\}, \{3, 5, 6\}, \{4, 5, 6\}]$ \\ \midrule
\end{tabular}
\caption{Label Partitions}
\label{table:lab_part}
\end{table}

We divided each dataset into train, validation, and test splits. All training occurs exclusively on the train split and all results are reported for performance on the test split. We use the validation split to construct each node's good-enough model space. We use a train/val/test split of $50000$/$5000$/$5000$ for MNIST and CIFAR-10. For HAM, we use a $80$/$10$/$10$ percentage split (since no conventional train/test partitioning exists). 

\subsection{Convex Model Training}\label{appen:convex}
Our convex model consists of a simple logistic regression classifier. We train with Adam, a learning rate of $0.001$, and a batch size of $32$. We terminate training when training accuracy converges.  

\subsection{Non-Convex Model Training}\label{appen:nonconvex}
Our non-convex model consists of a simple two layer feedforward neural network. For MNIST and HAM, we fix the hidden layer size to $50$ neurons. For CIFAR-10, we fix the hidden layer size to $100$ neurons. We apply dropout \cite{srivastava2014dropout} with a rate of $0.5$ to the hidden layer. We train with Adam, a learning rate of $0.001$, and a batch size of $32$. We terminate training when training accuracy converges.  

\section{Additional Results}

\subsection{Convex Results}\label{appen:convex_results}

Table \ref{table:full_convex} presents results for convex logistic regression over different values of $K$. We use $\epsilon = 0.40$ for MNIST, $\epsilon = 0.20$ for HAM, and $\epsilon = 0.20$ for CIFAR-10 (for $ K =5$ nodes). 

At each $K$, we compared the performance of modelling the good-enough model space as an ellipsoid or a $\mathbb{R}^d$ ball. We report the result corresponding to the best method. We found that using $\mathbb{R}^d$ balls as the good-enough model space ($H_k$) resulted in aggregate models almost exactly equivalent to the parameter average of all local models. In general, we observe that the performance of \gems and the baseline decreases as $K$ increases. However, the fine-tuned \gems performance stays relatively constant. 

 \begin{table}[h]
  \renewcommand{\arraystretch}{1.4}
    \caption{Convex Results}
    \label{table:full_convex}
    \begin{tabular}{lc|c|ccc|c}
    \hlinewd{1pt}
    \bf Dataset & $K$ & \bf Global & \bf Local &\bf Averaged  &\bf \gems &\bf \gems Tuned  \\ \hlinewd{0.5pt}
    MNIST  & 2& 0.926 (0.001)	&0.481 (0.027)&	\bf 0.780 (0.015)&		\bf 0.780 (0.015)&	0.889 (0.003)   \\ \hlinewd{0.5pt}
    MNIST  & 3& 0.926 (0.001) 	&0.325 (0.042)&	\bf 0.705 (0.013)&		0.647 (0.033)&	0.878 (0.004)   \\ \hlinewd{0.5pt}
    MNIST & 5 & 0.926 (0.001)	&0.198 (0.010)&	0.444 (0.028)&	\bf 0.456 (0.020)&	0.877 (0.005)     \\ \hlinewd{1.0pt}
    CIFAR-10 & 2 & 0.597 (0.006)	&\bf 0.385 (0.025)&	0.253 (0.027)&		0.234 (0.017)&	0.494 (0.009)  \\ \hlinewd{0.5pt}
    CIFAR-10 & 3 & 0.597 (0.006)	&0.272 (0.064)&	0.203 (0.022)&	\bf 0.300 (0.040)&	0.495 (0.011)  \\ \hlinewd{0.5pt}
    CIFAR-10 & 5 & 0.597 (0.006)	&0.178 (0.010)&	0.154 (0.024)&	\bf 0.210 (0.020)&	0.499 (0.016)  \\ \hlinewd{1.0pt}
    HAM   & 2 & 0.559 (0.002)	&0.344 (0.018)&	\bf 0.400 (0.020)&		0.353 (0.011)	&0.491 (0.006)   \\ \hlinewd{0.5pt}
    HAM   & 3 & 0.559 (0.002)	&0.269 (0.057)&	0.252 (0.043)&\bf	0.359 (0.045)&	0.523 (0.009)   \\ \hlinewd{0.5pt}
    HAM   & 5 & 0.559 (0.002)	&0.237 (0.050)&	0.185 (0.002)& \bf 0.348 (0.048)&	0.530 (0.009)   \\ \hlinewd{0.75pt}
    \end{tabular}
\end{table}

\subsection{Non-Convex Results}\label{appen:non_convex_results}
Table \ref{table:dnn_mnist} presents the non-convex results (two layer neural network) for MNIST. We use $\epsilon = 0.7$ for the final layer, and let $\epsilon_j$ denote the deviation allowed for  hidden neurons (specified in Eq \ref{eq:neuroneval}).

 \begin{table}[h]
  \renewcommand{\arraystretch}{1.4}
    \caption{MNIST Results (Neural Network)}
    \label{table:dnn_mnist}
    \begin{tabular}{ccc|c|ccc|c}
    \hlinewd{1pt}
    \bf $K$ & $\epsilon_\text{hidden}$  & $m_\epsilon$  & \bf Global    & \bf Local     &\bf Averaged      &\bf \gems       &\bf \gems Tuned     \\ \hlinewd{0.5pt}
         2  & 0.01  & 1   & 0.965 (0.001) & 0.492 (0.024) & 0.641 (0.058) &  \bf 0.766 (0.083) & 0.888 (0.004)     \\\hlinewd{0.5pt}
         3  & 1.0   & 100 & 0.965 (0.001) & 0.329 (0.043) & 0.422 (0.038) &  \bf 0.754 (0.024) & 0.926 (0.006)     \\\hlinewd{0.5pt}
         5  & 1.0   & 100  & 0.965 (0.001)  & 0.199 (0.010) & 0.259 (0.039) & \bf 0.439 (0.044) & 0.886 (0.007)     \\ \hlinewd{0.75pt}
\end{tabular}
\end{table}

Table \ref{table:dnn_cifar10} presents the non-convex results for CIFAR-10. We use $\epsilon = 0.2$ for the final layer. 

\begin{table}[h]
\renewcommand{\arraystretch}{1.4}
    \caption{CIFAR-10 Results (Neural Network)}
    \label{table:dnn_cifar10}
    \begin{tabular}{ccc|c|ccc|c}
    \hlinewd{1.0pt}
    \bf $K$ & $\epsilon_j$  & $m_\epsilon$  & \bf Global    & \bf Local     &\bf Averaged      &\bf \gems       &\bf \gems Tuned     \\ \hlinewd{0.5pt}
        2 & 0.1 & 1.0 &0.651 (0.004)  & \bf 0.405 (0.019) & 0.192 (0.026) &  0.335 (0.041) & 0.568 (0.007)    \\ \hlinewd{0.5pt}
        3 & 0.3 & 150 & 0.651 (0.004)  & 0.284 (0.061) & 0.163 (0.029) & \bf 0.333 (0.059) & 0.538 (0.009)    \\ \hlinewd{0.5pt}
        5 & 0.3 & 200 & 0.651 (0.004)  & 0.183 (0.009) & 0.128 (0.023) & \bf 0.223 (0.011) & 0.502 (0.011)    \\ \hlinewd{0.75pt}
\end{tabular}
\end{table}
Table \ref{table:dnn_ham} presents the non-convex results for CIFAR-10. We use $\epsilon = 0.25$ for the final layer. 

 \begin{table}[h]
 \renewcommand{\arraystretch}{1.4}
    \caption{HAM Results (Neural Network)}
    \label{table:dnn_ham}
    \begin{tabular}{ccc|c|ccc|c}
    \hlinewd{1.0pt}
    \bf $K$ & $\epsilon_j$  & $m_\epsilon$ & \bf Global    & \bf Local     &\bf Averaged      &\bf \gems       &\bf \gems Tuned     \\ \hlinewd{0.5pt}
        2 & 0.01 & 1.0 & 0.606 (0.002)  & 0.354 (0.022) & 0.273 (0.032) & \bf 0.399 (0.039) & 0.539 (0.008)     \\ \hlinewd{0.5pt}
        3 & 0.07 & 100 & 0.606 (0.002)  & \bf 0.271 (0.061) & 0.195 (0.042) &  0.269 (0.089) & 0.525 (0.014)     \\ \hlinewd{0.5pt}
        5 & 0.07 & 100 & 0.606 (0.002)  & \bf 0.250 (0.048)	&0.153 (0.018)&	0.190 (0.002)&	0.544 (0.017)    \\ \hlinewd{0.75pt}
\end{tabular}
\end{table}

In general, we observe that the baselines and \gems degrade as $K$ increases. Fine-tuning delivers a significant improvement, but is less consistent as $K$ varies.   

\subsection{Ensemble Results}\label{appen:ensemble}
We present full results for a comparison between ensemble methods and \gems. These results illustrate the modularity of \gems: by adjusting $\epsilon_j$ and $m_\epsilon$, the operator can tradeoff performance and model size. They also demonstrate that \gems is able to outperform ensembles, despite requiring far fewer parameters. For ease of clarity, we describe the model size in terms of the number of hidden neurons. For ensembles, we sum the hidden neurons across all ensemble members. All results are averaged over 5 trials, with standard deviations reported.

\begin{table}[h]
 \centering
 \caption{Model Size Results (MNIST, $K= 5$)}
    \label{table:mnist_model_size}
\begin{tabular}{lcc}
\toprule
\bf Method                                              & \bf Accuracy      & \bf \# hidden neurons \\ \midrule
Tuned \gems ($m_\epsilon = 75$, $\epsilon_j = 0.5$) & 0.872 (0.007) & 74.00 (0.00)          \\ \midrule
Tuned \gems ($m_\epsilon = 100$, $\epsilon_j = 1.0$) & 0.886 (0.007) & 99.00 (0.00)            \\ \midrule
Tuned \gems ($m_\epsilon = 50$, $\epsilon_j = 1.0$) & 0.862 (0.009) & 49.0 (0.00)            \\ \midrule
Tuned \gems ($m_\epsilon = 75$, $\epsilon_j = 1.0$) & 0.867 (0.008) & 79.00 (0.00)           \\ \midrule
Ensemble                                            & 0.210 (0.006) & 250.00 (0.0)            \\ \bottomrule
\end{tabular}
\end{table}

\begin{table}[h]
 \centering
 \caption{Model Size Results (CIFAR-10, $K= 5$)}
    \label{table:cifar10_model_size}
\begin{tabular}{lcc}
\toprule
\bf Method                                              & \bf Accuracy      & \bf \# hidden neurons \\ \midrule
Tuned \gems ($m_\epsilon = 150$, $\epsilon_j = 0.7$) & 0.454 (0.018) & 163.40 (1.20)            \\ \midrule
Tuned \gems ($m_\epsilon = 150$, $\epsilon_j = 0.5$) & 0.492 (0.012) & 246.20 (8.93)            \\ \midrule
Tuned \gems ($m_\epsilon = 200$, $\epsilon_j = 0.3$) & 0.502 (0.011) & 379.60 (6.68)            \\ \midrule
Tuned \gems ($m_\epsilon = 100$, $\epsilon_j = 0.3$) & 0.501 (0.011) & 386.00 (18.76)           \\ \midrule
Ensemble                                            & 0.194 (0.005) & 500.00 (0.0)            \\ \bottomrule
\end{tabular}
\end{table}

\begin{table}[h]
\centering
 \caption{Model Size Results (HAM, $K= 5$)}
    \label{table:ham_model_size}
\begin{tabular}{lcc} 
\toprule
\bf Method                                           & \bf Accuracy      & \bf Num hidden    \\ \midrule
Tuned \gems ($m_\epsilon = 75$, $\epsilon_j = 0.07$)  & 0.543 (0.015)&	103.80 (8.18)  \\ \midrule
Tuned \gems ($m_\epsilon = 100$, $\epsilon_j = 0.07$) & 0.544 (0.017)&	161.60 (7.23) \\ \midrule
Tuned \gems ($m_\epsilon = 50$, $\epsilon_j = 0.07$)  & 0.529 (0.013)&	 90.50 (3.77) \\ \midrule
Ensemble                                             & 0.245 (0.010) & 250          \\ \bottomrule
\end{tabular}
\end{table}

Both $m_\epsilon$ and $\epsilon_j$ loosely control the size of the hidden layer. $m_\epsilon$ effectively lower bounds the number of hidden units in the \gems model, and increase $m_\epsilon$ increases the size of the model\footnote{In some cases, $k-$means produces an empty cluster which contains no neurons}. Similarly, $\epsilon_j$ controls the amount of loss tolerated when constructing the good-enough space for each neuron. Increasing $\epsilon_j$ increases the likelihood of finding an intersection, thereby reducing the size of the hidden layer. 

\subsection{Intersection Analysis}\label{appen:intersect}
\begin{figure}[h]
    \centering
    \includegraphics{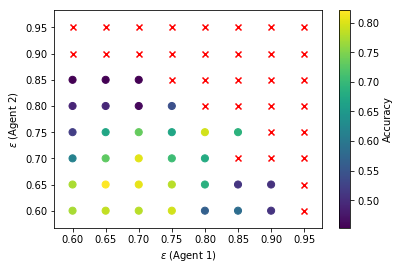}
    \caption{The x-axis corresponds to different settings of $\epsilon$ for the first node, and the y-axis corresponds to different settings of $\epsilon$ for the second node. Red crosses denote values where \gems failed to find an intersection. The color of the circular markers denotes the accuracy of the intersected model.}
    \label{fig:intersect_plot}
\end{figure}

We notice that in order for \gems to find an intersection, we have to set $\epsilon$ conservatively. We consider the convex MNIST case ($K = 2$) with $\mathbb{R}^d$ ball good-enough model spaces, and perform a grid search over different values of $\epsilon$ for each node. We illustrate these results in Figure \ref{fig:intersect_plot}. The X-axis corresponds to an $\epsilon$ value for node $1$, and the Y-axis corresponds to an $\epsilon$ value for node $2$. Markers denoted by a red cross indicate that no intersection was found at the corresponding $\epsilon$ settings. Markers denoted by a filled-inn circle indicate that a model in the intersection $h_G$ was identified, and the shade of the circle denotes the accuracy of $h_G$ on the global test data (with no tuning). We identify several trends. 

First, setting epsilon aggressively (i.e. at a higher value) for both nodes results in no intersection. This is most likely explained by the coarseness of $\mathbb{R}^d$ balls as an approximation to the good-enough model space. Because Algorithm 2 approximates the largest $\mathbb{R}^d$ ball inscribed in the good-enough space, higher values of $\epsilon$ will produce smaller $\mathbb{R}^d$ balls, reducing the likelihood of an intersection. We believe the lack of intersections at higher values of $\epsilon$ can be addressed with more sophisticated representations of the good-enough model space, which better capture the topology of each node's loss surface. 

Second, setting epsilon aggressively for one node (and conservatively for the other) does result in \gems  identifying a model, albeit with poorer performance. In these cases, Algorithm 2 will learn a 'tight' ball (i.e. small radius) for the node with the higher $\epsilon$ and a 'loose' ball for the node with the smaller $\epsilon$. Thus, the model identified by \gems will be much closer to the local model of the node with the higher value of $\epsilon$. As a result, the performance of the \gems model significantly degrades, and more closely resembles the performance of one of the local models. 

We observe the best performance when each node's $\epsilon$ is approximately equivalent, and set relatively conservatively. This ensures that 1) an intersection is identified, and 2) the model learned via \gems does not strongly favor one node. Interestingly, the performance of the \gems model can often exceed $\epsilon$. This suggests inefficiencies in our approximation of the good-enough model space that could easily be improved with more sophisticated representations.

\end{document}